\documentclass[pdflatex,sn-mathphys-num]{sn-jnl}

\usepackage{graphicx}%
\usepackage{multirow}%
\usepackage{amsmath,amssymb,amsfonts}%
\usepackage{amsthm}%
\usepackage{mathrsfs}%
\usepackage[title]{appendix}%
\usepackage{xcolor}%
\usepackage{textcomp}%
\usepackage{manyfoot}%
\usepackage{booktabs}%
\usepackage{algorithm}%
\usepackage{algorithmicx}%
\usepackage{algpseudocode}%
\usepackage{listings}%
\usepackage{tikz}%
\usepackage{graphicx}%
\usepackage{subcaption}

\theoremstyle{plain}
\theoremstyle{thmstyletwo}%

\theoremstyle{thmstylethree}%

\begin{document}

\title[Article Title]{Vision Meets Language: A RAG-Augmented YOLOv8 Framework for Coffee Disease Diagnosis and Farmer Assistance}


\author*[1]{\fnm{Semanto} \sur{Mondal}}\email{semanto.mondal@unina.it}

\affil*[1]{ \orgname{University of Naples Federico II}, \orgaddress{\street{Via Cinthia 21}, \city{Italy}, \postcode{80126}, \state{Naples}, \country{Italy}}}


\abstract{As a social being, we have an intimate bond with the environment. A plethora of things in human life, such as lifestyle, health, and food are dependent on the environment and agriculture. It comes under our responsibility to support the environment as well as agriculture, which can ensure us a fruitful and healthy life. However, traditional farming practices often result in inefficient resource use and environmental challenges. To address these issues, precision agriculture has emerged as a promising approach that leverages advanced technologies to optimise agricultural processes. Nowadays, due to the advancement of technology, many different artificial intelligence technologies have been developed to tackle different issues related to agriculture and the environment. 

Reducing pesticide usage presents a significant challenge in contemporary agriculture; however, a pragmatic solution may be found in the development of a hybrid approach. In this work, a hybrid approach is proposed that combines the three different potential fields of model AI: object detection, large language model (LLM), and  Retrieval-Augmented Generation (RAG). In this novel framework, we have tried to combine the vision and language models to work together to identify potential diseases in the tree leaf. This study introduces a novel AI-based precision agriculture system that uses Retrieval Augmented Generation (RAG) to provide context-aware diagnoses and natural language processing (NLP) and YOLOv8 for crop disease detection. The system aims to tackle major issues with large language models (LLMs), especially hallucinations and allows for adaptive treatment plans and real-time disease detection. The system provides an easy-to-use interface to the farmers, which they can use to detect the different diseases related to coffee leaves by just submitting the image of the affected leaf the model will detect the diseases as well as suggest potential remediation methodologies which aim to lower the use of pesticides, preserving livelihoods, and encouraging environmentally friendly methods.  With an emphasis on scalability, dependability, and user-friendliness, the project intends to improve RAG-integrated object detection systems for wider agricultural applications in the future. The global movement for more intelligent, sustainable food production systems is aided by this work.
}




\maketitle

\section{Introduction}\label{sec1}

AI can be applied across various aspects of agriculture, including irrigation management, environmental monitoring, agricultural surveillance, and anomaly detection. Advanced techniques like remote sensing have gained significant traction in agriculture, particularly for tasks such as crop detection, localisation, and object identification. Moreover, drone technology integrated with AI has opened new possibilities in precision farming. For instance, concepts like AI-based Learned Inertial Odometry, initially developed for autonomous drone racing, can be adapted to monitor fields, manage agricultural environments, and identify anomalies effectively.

Pesticides are becoming more prevalent in agriculture to protect crops and increase crop yields. It can also minimise the yield loss of up to 40\% globally which occurs due to pests and diseases \cite{FAO}. Farmers are utilising different spraying techniques to utilise pesticides which can lead to spray drift and losses of pesticides into surface waters and the environment. On the other hand, pesticide application also results in residues, which are an acute food-related concern. More than 45\% of food products are found to contain traces of pesticides \cite{EFSA}. According to a recent analysis, the extensive usage of pesticides is also imposing tremendous threats to different species Thus, “1,000,000 species (are) threatened with extinction”, which is in large part attributed to unsustainable land use and pollution, which includes the excessive use of pesticides \cite{EIP}. This also imposes an unavoidable threat to the balance of the ecosystem, which might result in ecological imbalance if not treated on an early basis.

Object detection and localisation are very common in terms of leaf disease detection and classification. Among the available object detection models YOLO (You Only Look Once) is the most affection one, which is initially trained on a large COCO dataset. In general, YOLO is an object detection method that processes an entire image in one go, simultaneously predicting bounding boxes and class probabilities. It uses anchor boxes to effectively detect objects of various sizes and shapes. With its grid-based design and powerful convolutional neural network backbone, YOLO delivers fast, real-time performance while maintaining high accuracy across a wide range of applications. The original YOLOv1, YOLOv2 (YOLO9000), YOLOv3, YOLOv4, and YOLOv5 were the first in the line of YOLO
object detection models. The most recent models in the line are YOLOv6, YOLOv7, YOLOv8, YOLOv9, YOLOv10, YOLOv11 and YOLOv12. Every iteration has sought to improve speed, precision, and adaptability for a range of uses, such as plant disease detection \cite{Ultralytics}. These object detection models can detect potential issues but do not have reasoning ability, such as telling the reason for the diseases or going beyond detection to generate some human-understandable texts.

Large Language Models (LLMs) are advanced artificial intelligence systems trained on vast amounts of text data to understand and generate human-like language. The underlying working principle is based on the transformer architecture. LLMs like GPT, BERT, and their successors have significantly improved the state of AI in language understanding and become experts in various tasks such as translation, summarisation, question-answering, and content generation. Unfortunately, LLMs suffer from hallucination, which is nothing but making up unrealistic answers or irrelevant answers. 

In this paper, we have tried to overcome the limitations of both the vision and language model by implementing a RAG-Augmented YOLOv8 Framework. We aim to add additional information to the output of the YOLO model to make the output interpretable using the Llama-3 and mitigate the hallucination issue of LLM by introducing a RAG based chatbot where there is a knowledge base which is used as a symbolic component and give information about the potential diseases and remedies so that the LLM can generate a fine-grained text output using both the output of YOLO and information retrieval from the external knowledge base. Overall, the proposed method can be treated as a neurosymbolic approach where yolo and LLM are the neural components, and the external knowledge-based is the symbolic component, which provides context to the LLM to generate guided feedback.

\section{Related Work}\label{sec2}
The integration of deep learning techniques into agriculture has significantly advanced plant disease detection, offering rapid and accurate diagnostics that surpass traditional manual methods. Among these techniques, the You Only Look Once (YOLO) family of models has garnered attention for their real-time object detection capabilities. YOLOv8, the latest iteration, has been applied to various crops, demonstrating high accuracy and efficiency.

For instance, a study on Bangladeshi crops utilized YOLOv8 \cite{bangladesh_yolo} to detect leaf diseases in rice, corn, wheat, potato, and tomato, achieving a mean Average Precision (mAP) of 98\% and an F1 score of 97\%. Similarly, research on tomato disease detection \cite{tomato} in Saudi Arabia employed YOLOv8, resulting in an overall accuracy of 66.67\%, highlighting the model's potential in real-world agricultural settings.

Enhancements to the YOLOv8 architecture have further improved its performance. The SerpensGate-YOLOv8 \cite{SerpensGate} model introduced modifications such as Dynamic Snake Convolution and Super Token Attention mechanisms, leading to a 3.3\% improvement in mAP@0.5 over the baseline YOLOv8. Additionally, the Pyramid-YOLOv8 model \cite{pyramidyolo}, designed for rice leaf blast detection, incorporated multi-scale feature fusion and attention modules, achieving higher detection accuracy and robustness in complex field environments.

The FarmTalk-Nexus \cite{nexus} platform integrates YOLOv8 for plant disease detection with GPT-3.5 Turbo using a Retrieval-Augmented Generation (RAG) framework. This system provides farmers with real-time, context-specific remediation strategies through an interactive interface, demonstrating the practical application of combining computer vision and LLMs in agriculture. Yu and Zutty (2025) \cite{llmguided} introduced the LLM-Guided Evolution (LLM-GE) framework, which leverages LLMs to autonomously optimise YOLO architectures for object detection tasks. By incorporating the "Evolution of Thought" (EoT) technique, the system iteratively refines YOLO models, achieving a notable increase in Mean Average Precision (mAP) from 92.5\% to 94.5\% on the KITTI dataset. Roumeliotis et al. \cite{roumeliotis2025plantdiseasedetectionmultimodal} explored the integration of multimodal LLMs, specifically GPT-4o, with Convolutional Neural Networks (CNNs) for automated plant disease classification using leaf imagery. Their study demonstrated that fine-tuned GPT-4o models slightly outperformed traditional CNNs like ResNet-50, achieving up to 98.12\% accuracy on apple leaf images. 

The integration of Retrieval-Augmented Generation (RAG) frameworks with Large Language Models (LLMs) has significantly advanced the field of knowledge-intensive tasks, offering improved reasoning capabilities and accessibility. Feng et al. \cite{rag1} introduced a Retrieval-Generation Synergy framework that iteratively combines retrieval and generation processes, enhancing the cognitive reasoning capacity of LLMs. This approach demonstrated superior performance across multiple question-answering datasets, outperforming traditional retrieval-only or generation-only methods. In the legal domain, Louis et al. \cite{rag2} developed an interpretable long-form legal question-answering system utilising a "retrieve-then-read" pipeline. By introducing the Long-form Legal Question Answering (LLeQA) dataset, comprising 1,868 expert-annotated legal questions in French, the study showcased the potential of RAG frameworks in generating syntactically correct and contextually relevant responses to complex legal queries. Addressing the needs of non-specialist users, Dodgson et al. \cite{rag3} explored methods to enhance LLM performance through fine-tuning, RAG, and soft-prompting. Their study revealed that RAG approaches outperformed both unmodified and fine-tuned GPT-3.5 models, especially when combined with basic soft prompts. This underscores the effectiveness of RAG in improving LLM outputs, even with limited datasets and technical expertise.

The use of large language models (LLMs) in agriculture has created new opportunities for contextual information and plant disease treatments beyond visual detection. LLMs and domain-specific knowledge bases are combined in Retrieval-Augmented Generation (RAG) frameworks, which have demonstrated promise in providing precise and contextually relevant answers. However, research on the combination of conversational AI and real-time visual detection is still in its infancy. Although separate elements such as RAG for information retrieval and YOLOv8 for detection have been investigated, their combined use in a single system for plant disease diagnosis and treatment recommendation is comparatively new.

This is meant to be an all-in-one combination for end-users, the farmer, and agronomist, and not only highlights disease, but also offers a veritable treatable insight; such systems can provide treatments and information and use computer vision and natural language processing to make better agricultural decisions, which can affect crop quality and yield.

\section{RAG Based Chatbot}\label{sec3}
In general, chatbots are used for information retrieval. Traditional chatbots typically work based on some predefined rules as well as keyword matching. It’s kind of a set of if-else rules predefined by the users. When a user inputs a query, the chatbot searches for specific keywords or patterns in the query to identify the appropriate response. Traditional chatbots rely on a fixed knowledge base or database of predefined responses. The responses are manually inserted into the database by the developer. When a user inserts a query, the chatbot looks for the rules which are suitable based on the question. When it finds the question then it gives the answer which is hardcoded associated with that question. It doesn’t make any paraphrasing or can’t perform any generation. Nowadays, LLM-based chatbots are in the hype. LLM-based chatbots can be of two types. 

\textbf{LLM-Based Chatbots without RAG} Large Language Models (LLM) such as OpenAI, and Llama are trained with billions of parameters as well as with huge amounts of textual data. Some of these are open-source means that they can be used without any payment, and some are not. We can use the chatbot for our purpose using the API provided by the respected organisation of these LLMs. But here the problem is that when a user asks any question, it will directly answer from the data it has been trained on without considering any external knowledge base. It will work just like ChatGPT.  

\textbf{LLM-Based Chatbots with RAG} RAG stands for Retrieval-Augmented Generation. It has two main components: generation and retrieval. Unlike LLM-based chatbots without the RAG concept, here external data sources such as PDF, text, and database are used as a knowledge base along with the trained LLM model. So in this case, when any user asks for a query, it first looks for a similar type of text chunk in the external knowledge base, which is named retrieval. These text chunks are used as prompts to the LLM model. Based on the context and user query, the LLM model can create a more precise and creative answer, which can be referred to as generation. This is not possible with other types of chatbots. 

\begin{figure}[!htbp]
\centering
\includegraphics[width=\textwidth]{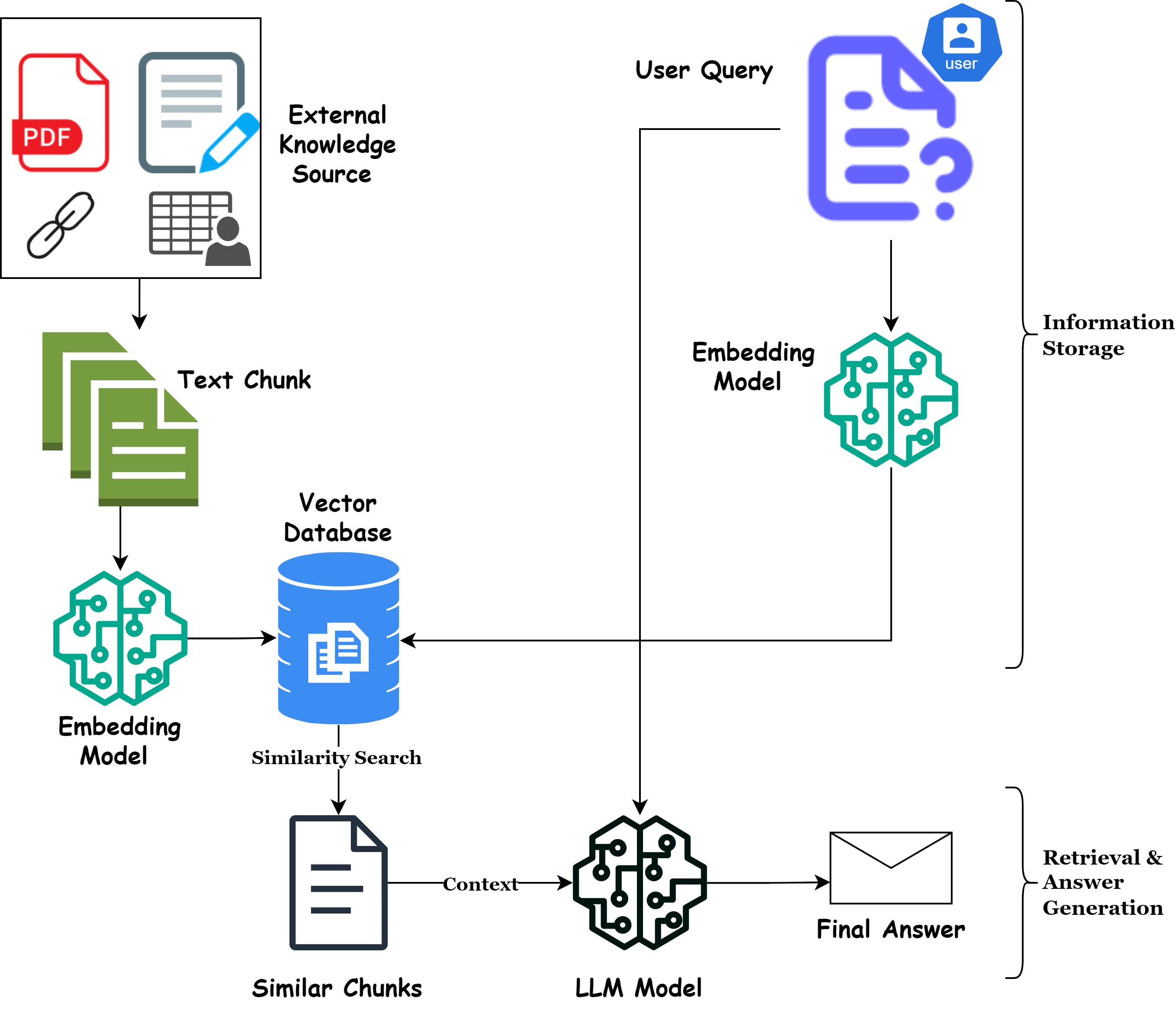}
\caption{High-Level Overview of the RAG-Based Chatbot}
\label{RAG Chatbot}
\end{figure}

The working of the RAG-based chatbot can be divided into three main parts. Figure \ref{RAG Chatbot} shows the workflow diagram of the RAG-based chatbot. 

\textbf{Information Storage} Varieties of documents can be used as external data sources. When any document is uploaded first using RecursiveCharacterTextSplitter, the text documents are split into chunks using overlapping or non-overlapping. Then, an embedding model is used to make an embedding vector of these text chunks to capture the semantic meaning. These embedding vectors are then stored in vector datasets using indexing for fast and precise information retrieval. 

\textbf{Information Retrieval} When a user inserts a textual query, using the same embedding model this text is converted into an embedding vector and passed to the vector database for information retrieval. Here the concept similarity search is used. Various techniques such as cosine similarity, L1 distance, or L2 distance, are used to retrieve similar contextual text chunks related to the user query. After retrieval, these chunks of documents are passed to the LLM model to generate the final answer. 

\textbf{Answer Generation} This is the final step of the chatbot. After information retrieval, the embedding vectors are passed to the LLM model. This is used as a prompt for the model. It can understand the context for which it needs to generate the answer. Finally, using the context, the generative LLM model generates the final output.

Figure \ref{RAG Chatbot} shows the overall workflow of the RAG based chatbot. 

\section{Methodology}\label{sec4}

\begin{figure}[!htbp]
\centering
\includegraphics[width=\textwidth]{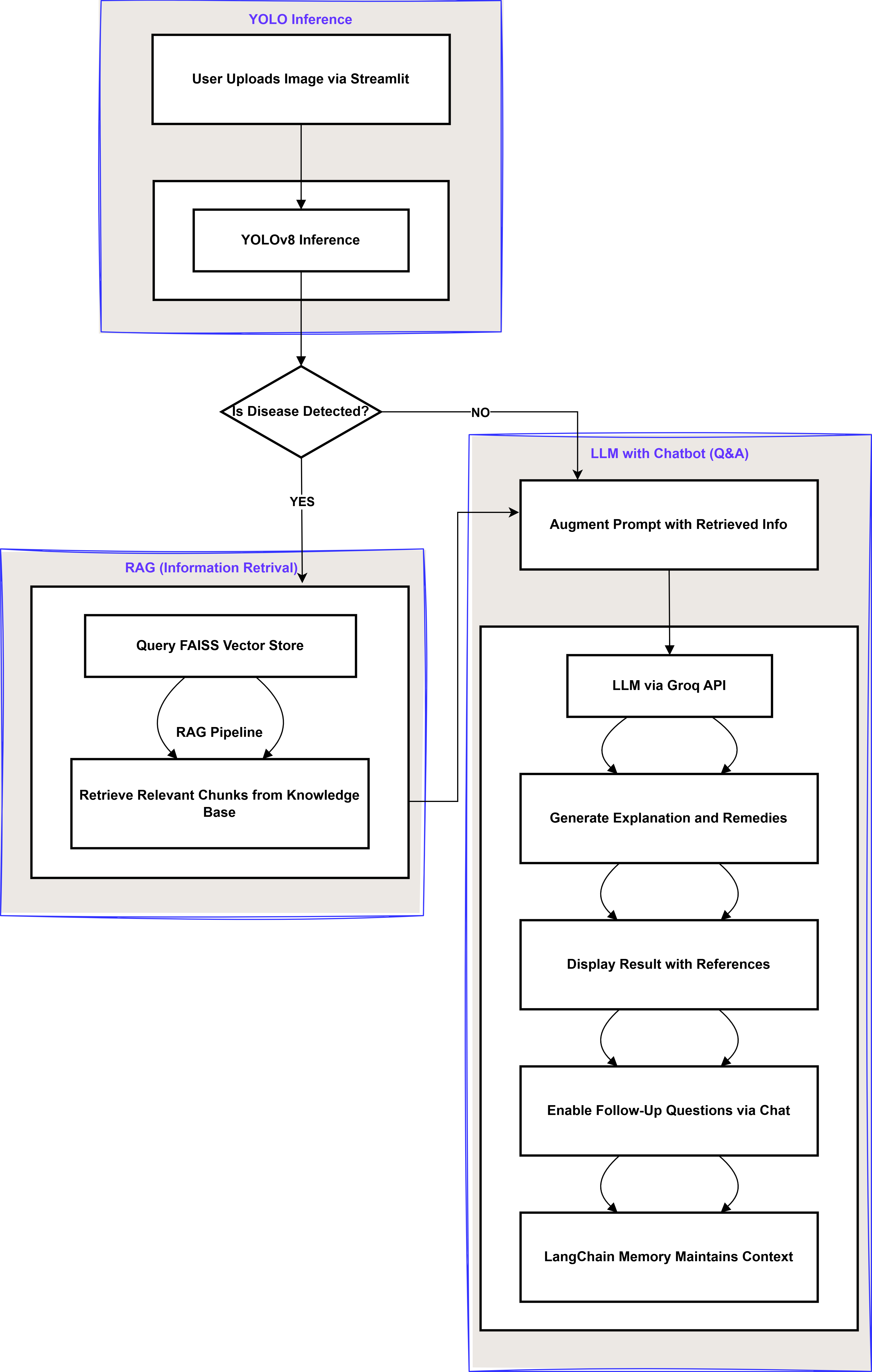}
\caption{Workflow of YOLOv8 + RAG + LLM-Based Coffee Leaf Disease Assistant}
\label{workflow}
\end{figure}

The hybrid methodologies are illustrated in Figure \ref{workflow} for coffee leaf
disease detection and remediation.

The overall workflow consists of three components as follows:
\subsection{YOLOV8 for Detecting Leaf Diseases}
The detection phase begins with the preparation of the dataset, where annotations and images are organised into training, validation, and test splits, defined within a data.yaml configuration file. We employed the YOLOv8-nano model (yolov8n.pt) from the Ultralytics framework due to its optimal balance between performance and computational efficiency. Given that the original dataset comprised high-resolution images sized at 2048×1024 pixels, which posed memory and processing constraints on the local setup, the input resolution was resized to 640×640 pixels. This not only accelerated the training process but also maintained sufficient detail for accurate disease detection.

The model was trained over 100 epochs using the default hyperparameters recommended by Ultralytics, including a batch size of 16, a learning rate of 0.01, momentum of 0.937, and weight decay of 0.0005, optimised using SGD. During training, the model leveraged various data augmentation strategies such as mosaic augmentation and horizontal flipping to increase robustness. Upon completion, the fine-tuned model weights were saved and used later for inference within the application pipeline.

When a user uploads an image through the Streamlit interface, it is passed directly to the trained YOLOv8 model for inference. The model outputs the predicted disease label(s) along with the corresponding bounding boxes, which are then visualised on the frontend. These predicted disease classes serve as input prompts to the language model in the subsequent stage.

\subsection{RAG for Information Retrieval}
To enable the system to generate accurate and contextually rich responses about detected plant diseases, a Retrieval-Augmented Generation (RAG) pipeline was implemented. This symbolic component plays a crucial role in grounding the Large Language Model (LLM) with domain-specific knowledge. We created a detailed document that serves as a knowledge base describing various coffee leaf diseases, symptoms, causes, and remedies. This document was then chunked and embedded using Hugging Face’s sentence-transformers embedding models and stored in a FAISS vector store for efficient similarity-based retrieval.

At inference time, once the YOLOv8 model identifies the disease, the resulting disease name is used as a query to the FAISS vector store. This triggers a retrieval of the top relevant passages from the knowledge base, which are subsequently appended to the user prompt. The augmented prompt—containing both the user’s question and relevant contextual information is then passed to the LLM for answer generation. This retrieval-then-read mechanism ensures that responses remain factually grounded and contextually relevant, improving the interpretability and precision of the system.

\subsection{LLM for Final Output Generation}
The final response generation is managed by a Large Language Model (LLM), integrated via the Groq API for low-latency and high-efficiency inference. The LLM receives a prompt that includes both the detected disease(s) and the associated retrieved content from the FAISS-powered knowledge base. This ensures that the model’s outputs are aligned with factual, curated information, rather than relying solely on pretrained weights.

The LLM processes this input to provide a comprehensive explanation of the detected disease, its likely causes, symptoms, and a list of actionable remedies. The interface also allows the user to engage in a conversational dialogue with the model—asking follow-up questions like “Why was this disease detected?” or “What if I apply the suggested remedy late?” The conversation is maintained using LangChain’s ConversationBufferWindowMemory, which retains recent interaction history to ensure context continuity. Additionally, for each answer generated, the system displays the source reference(s) of the retrieved information, thus enhancing the transparency and trustworthiness of the responses.

This modular integration: YOLOv8 for vision, RAG for retrieval, and LLM for generation is hosted on Streamlit for the end users so that users can directly use the web interface and interact with the underlying framework by uploading affected leaf images and natural language-based queries in a Q\&A manner.

\section{Dataset Description} \label{Data_Description}

\subsection{Dataset Acquisition}
\label{Data Acquisition}
In this work, we have used the publicly available BRACOL dataset from Kaggle, where BRACOL stands for Brazilian Arabica Coffee Leaf for Object Detection. Two different versions of the dataset are available. The first one is the original version \cite{original_dataset}. After that, the images are again annotated and rectified by a domain expert and generate the second version \cite{updated_dataset}. For this work, we have trained the YOLOV8 model using both versions of the dataset and observed that the best performance was obtained when the model was trained using the original version of the dataset.

\subsection{Dataset Overview}
\label{Detailed Description}
The BRACOL dataset comprises 1,747 images of Arabica coffee leaves, capturing both healthy specimens and those affected by various diseases. Initially, the dataset included 1,899 annotations across four disease categories: leaf miner (540), rust (621), phoma (464), and cercospora (274). However, to enhance the dataset's quality and comprehensiveness, a plant pathology expert conducted a thorough review and reannotation process. This meticulous effort involved correcting existing labels, refining bounding boxes, and identifying previously unmarked diseased regions. As a result, the total number of annotations increased significantly to 8,226, with the following distribution: rust (6,013), phoma (1,671), leaf miner (341), and cercospora (201).

\section{Experimental Settings}\label{Experimental settings}
The experiments were conducted on a local machine equipped with an NVIDIA RTX 4060 GPU featuring 8 GB of VRAM and 16 GB of system RAM, utilising CUDA version 11.8 to leverage GPU acceleration for training processes. For the object detection tasks, the Ultralytics YOLOv8 nano model (yolov8n.pt) was employed, trained over 100 epochs with an input image size of 640 pixels. Default training parameters were maintained, including a batch size of 16, an initial learning rate (lr0) of 0.01, momentum set at 0.937, and a weight decay of 0.0005.

For the development of the Large Language Model (LLM) and RAG components, APIs from Groq and Hugging Face were utilised. These platforms facilitated the integration of advanced language models and embedding techniques, enabling the system to process and perform inference on the complex textual data effectively.

\section{Results}\label{Result}

\subsection{Diseases Detection}
After training the YOLOv8n model on the curated and re-annotated BRACOL dataset, we evaluated the model performance on a separate validation set comprising 228 images with 402 annotated instances. The model was fine-tuned using 100 epochs with an input resolution of $640 \times 640$ on a local GPU-enabled machine (NVIDIA RTX 4060, 8GB VRAM). The final trained model file, \texttt{best.pt}, was used for inference.

Table~\ref{tab:detection_metrics} presents the class-wise and overall detection metrics including Precision, Recall, mAP@0.5, and mAP@0.5:0.95. The results demonstrate satisfactory performance, particularly for the \textit{Miner} and \textit{Phoma} disease classes. The model achieved an overall mean average precision (mAP@0.5) of 0.681 and mAP@0.5:0.95 of 0.454. Among the four disease types, the \textit{Miner} class showed the highest performance with mAP@0.5 reaching 0.894, indicating strong detection accuracy and localisation capability.

\begin{table}[h!]
\centering
\caption{Class-wise Detection Performance of YOLOv8n on Validation Set}
\label{tab:detection_metrics}
\begin{tabular}{lcccc}
\hline
\textbf{Class} & \textbf{Precision} & \textbf{Recall} & \textbf{mAP@0.5} & \textbf{mAP@0.5:0.95} \\
\hline
Cercospora & 0.546 & 0.630 & 0.575 & 0.329 \\
Miner & \textbf{0.823} & \textbf{0.849} & \textbf{0.894} & \textbf{0.650} \\
Phoma & 0.727 & 0.877 & 0.839 & 0.612 \\
Rust & 0.561 & 0.316 & 0.415 & 0.223 \\
\hline
\textbf{Overall} & 0.664 & 0.668 & 0.681 & 0.454 \\
\hline
\end{tabular}
\end{table}

It is evident that classes with a higher number of representative annotations (\textit{Miner} and \textit{Phoma}) yield better results, while detection of \textit{Rust} still poses challenges due to annotation imbalance and visual similarity to healthy leaf textures. Further data augmentation and class-balancing techniques may help improve detection in such underperforming categories.

\begin{figure}[h!]
    \centering
    \begin{subfigure}[b]{0.45\textwidth}
        \centering
        \includegraphics[width=\textwidth]{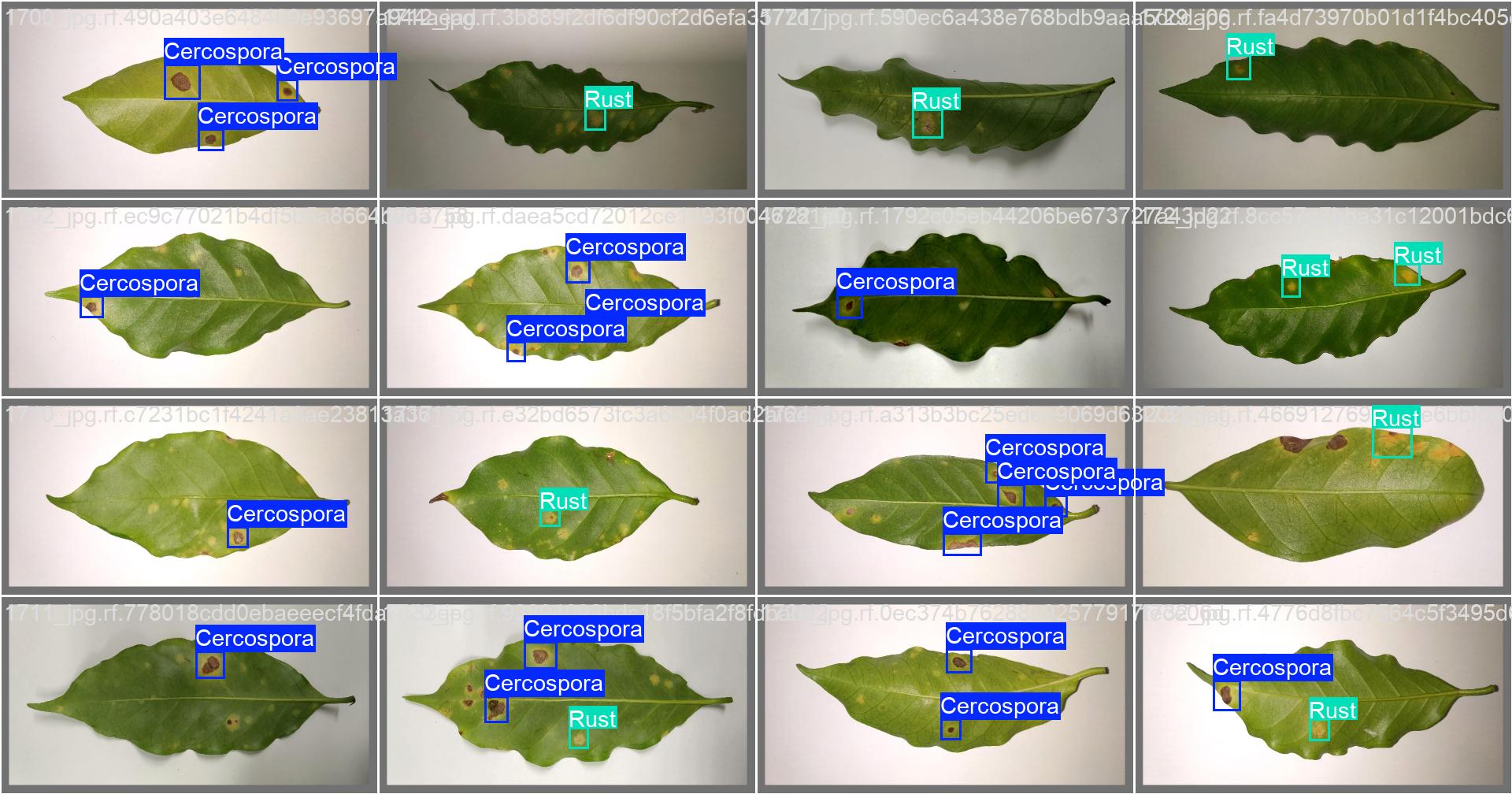}
        \caption{Ground Truth Labels}
        \label{fig:label_image}
    \end{subfigure}
    \hfill
    \begin{subfigure}[b]{0.45\textwidth}
        \centering
        \includegraphics[width=\textwidth]{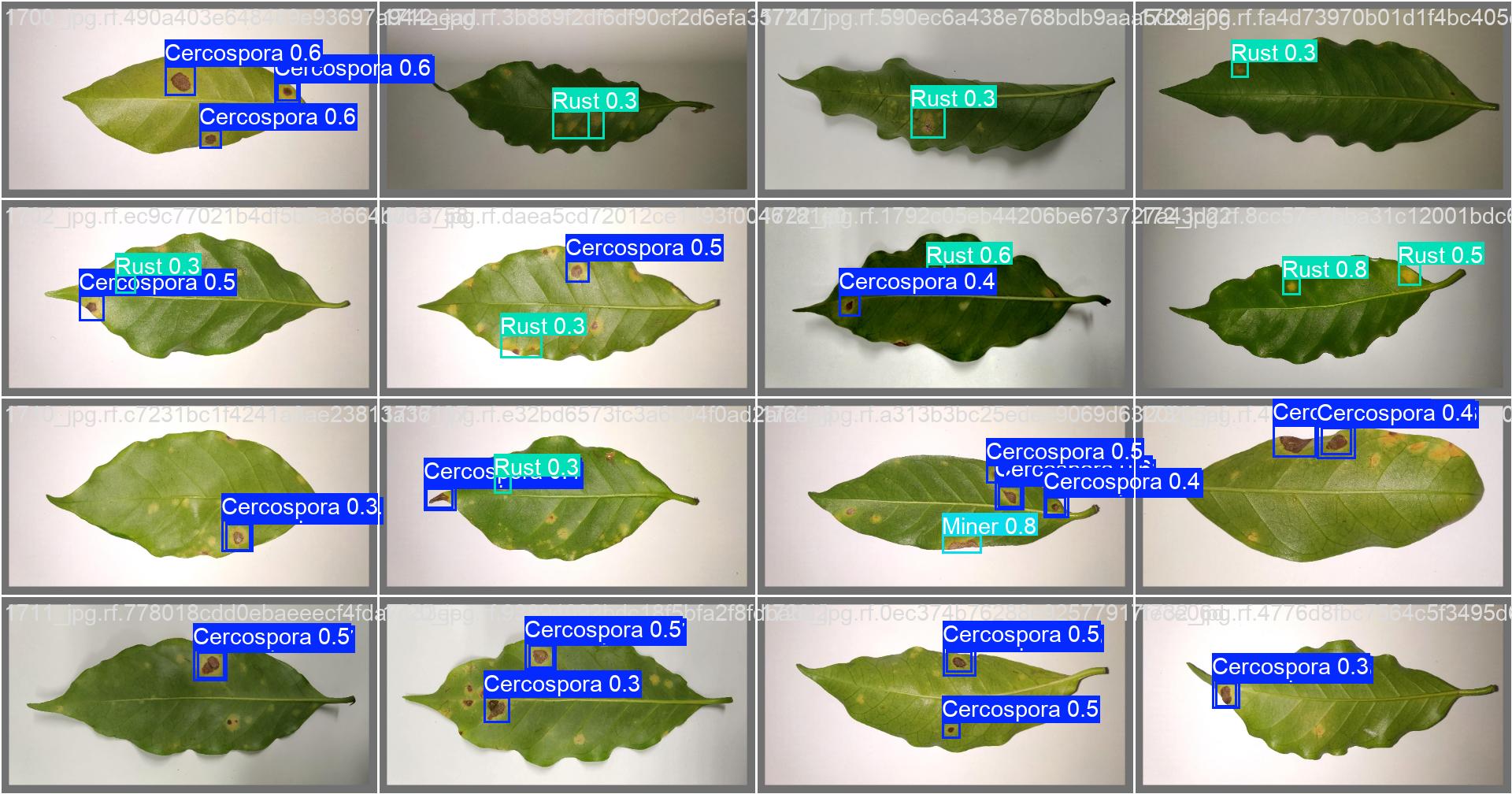}
        \caption{YOLOv8 Predictions}
        \label{fig:prediction_image}
    \end{subfigure}
    \caption{Comparison of annotated labels and YOLOv8 prediction results for a sample input image.}
    \label{fig:label_vs_prediction}
\end{figure}

\subsection{Remedy Generation and Follow-up Questions}

\begin{figure}[h!]
    \centering
    \begin{subfigure}[b]{0.45\textwidth}
        \includegraphics[width=\textwidth]{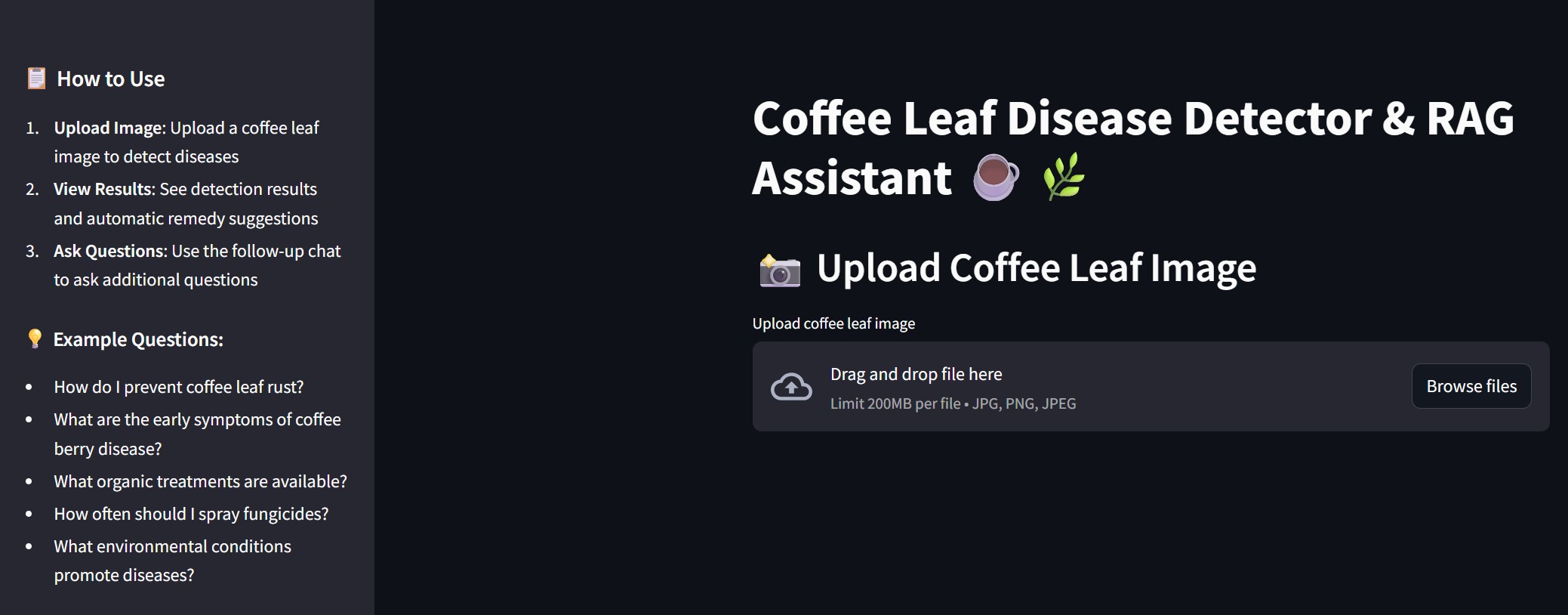}
        \caption{User interface for image upload and interaction}
        \label{fig:ui_interface}
    \end{subfigure}
    \hfill
    \begin{subfigure}[b]{0.45\textwidth}
        \includegraphics[width=\textwidth]{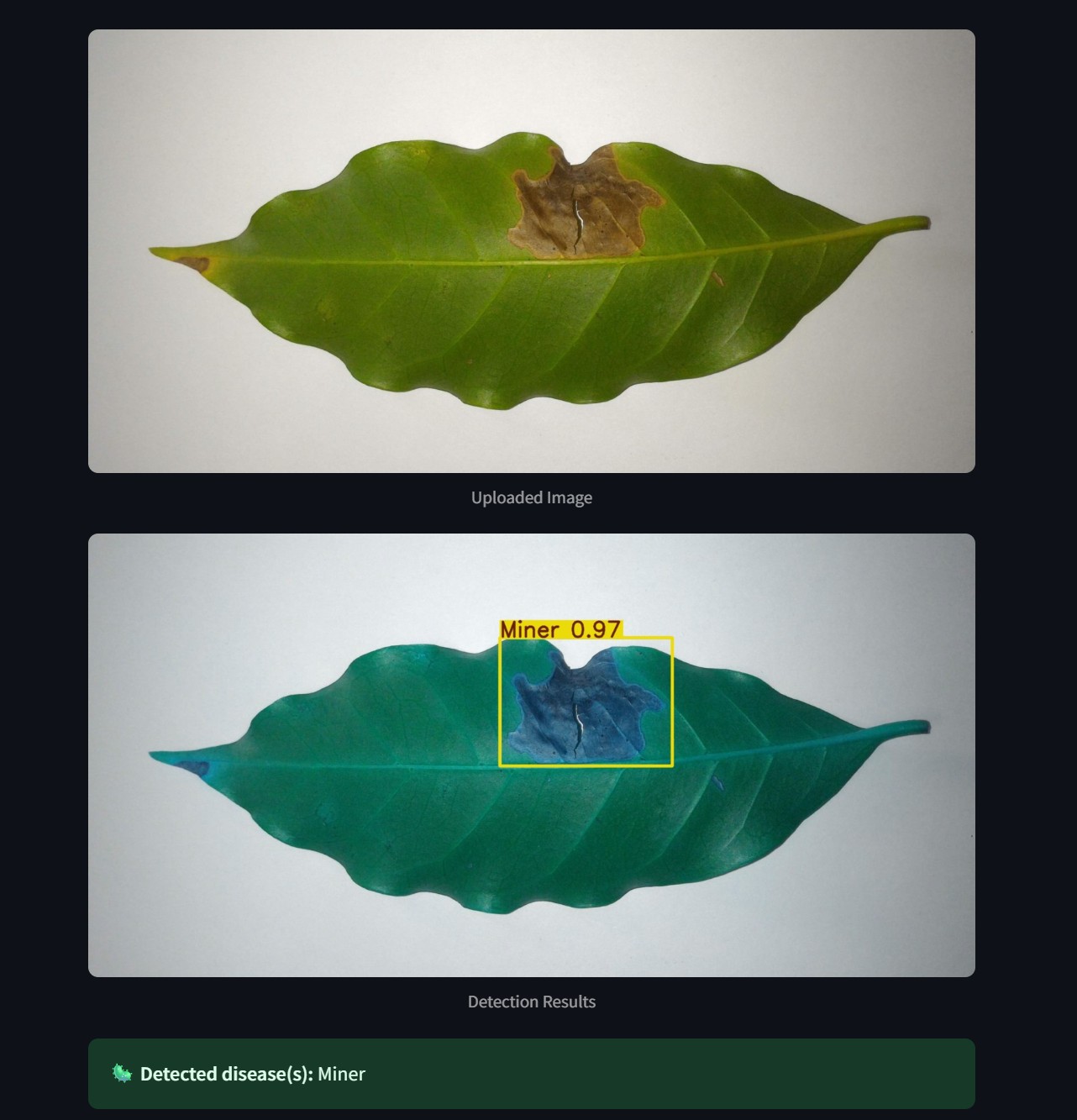}
        \caption{YOLOv8 detection output highlighting diseased regions}
        \label{fig:detection_result}
    \end{subfigure}
    \\
    \vspace{0.4cm}
    \begin{subfigure}[b]{0.45\textwidth}
        \includegraphics[width=\textwidth]{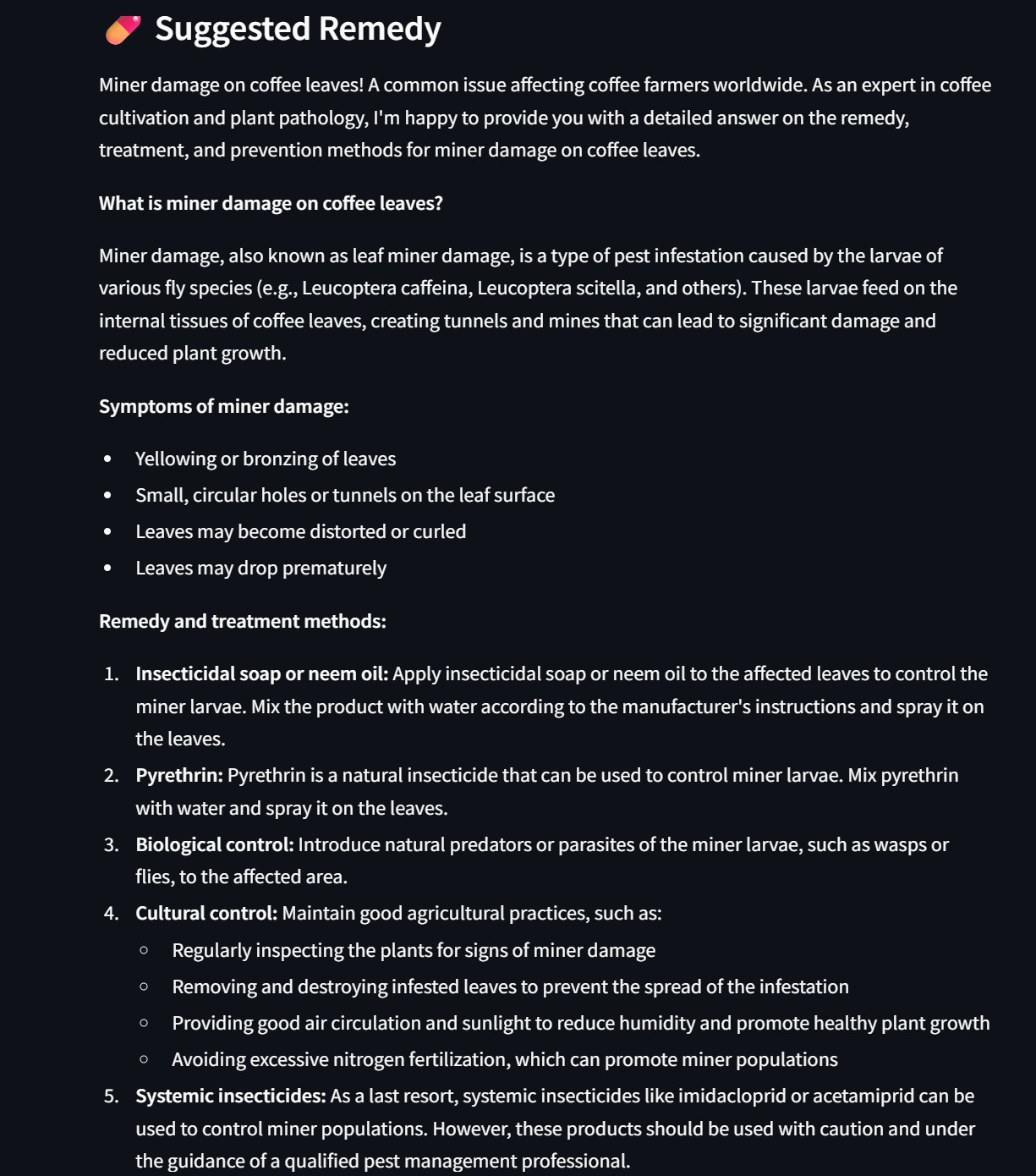}
        \caption{Generated remedy and disease explanation}
        \label{fig:remedy_output}
    \end{subfigure}
    \hfill
    \begin{subfigure}[b]{0.45\textwidth}
        \includegraphics[width=\textwidth]{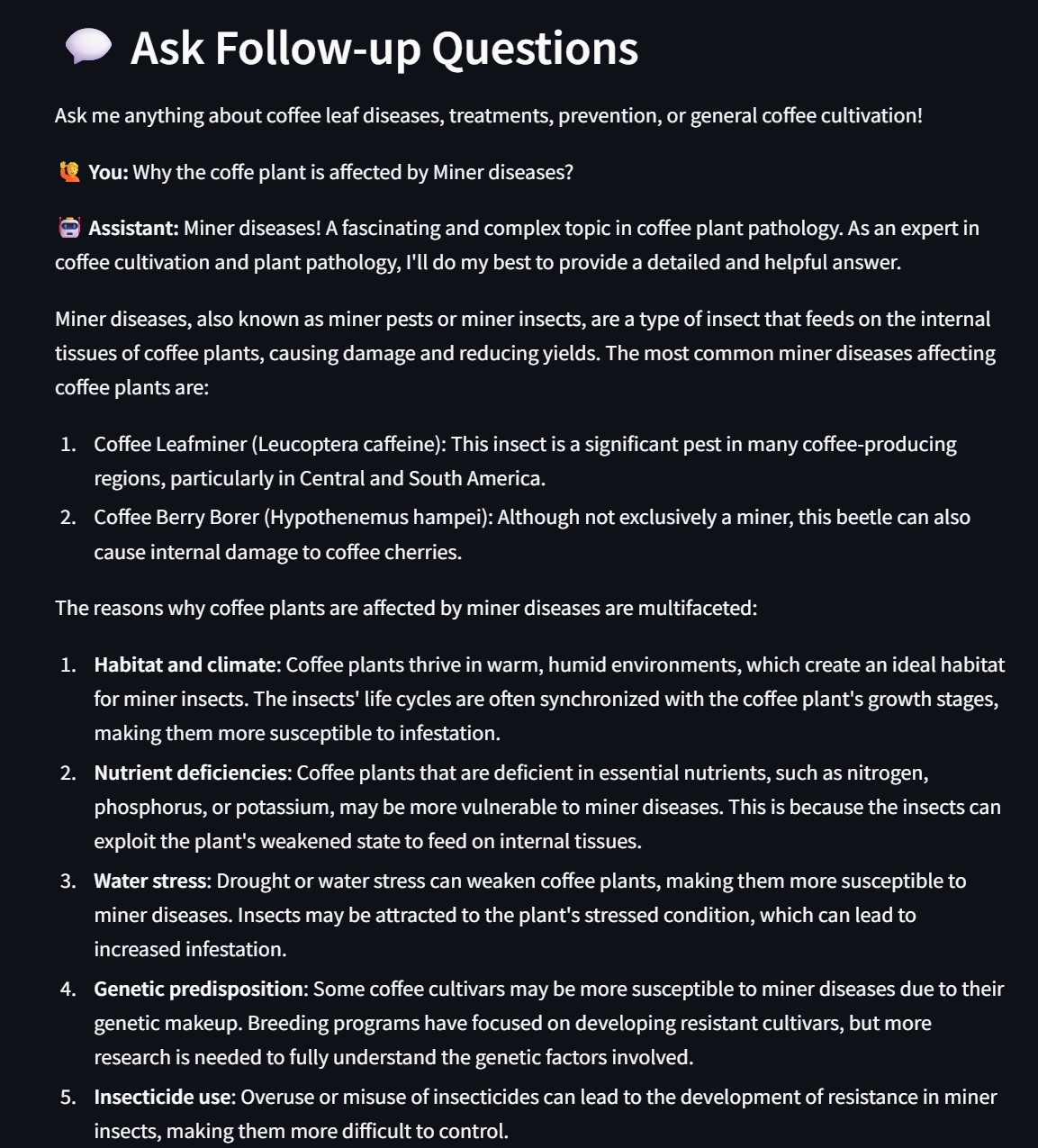}
        \caption{Follow-up question interface and response}
        \label{fig:followup_qa}
    \end{subfigure}
    \caption{End-to-end flow from image upload to disease remedy and conversational assistance.}
    \label{fig:remedy_workflow}
\end{figure}

After a leaf image is uploaded through the Streamlit-based user interface (Figure~\ref{fig:ui_interface}), the system first performs disease detection using the YOLOv8 model (Figure~\ref{fig:detection_result}). If a disease is detected, the predicted class label is used to form a query that is passed to the RAG (Retrieval-Augmented Generation) pipeline. This pipeline retrieves relevant contextual information from the FAISS vector store, which contains domain-specific knowledge extracted from agricultural documents. The retrieved content is used to augment the prompt sent to the LLM (via Groq API), which generates a detailed description of the disease and its suggested remedies (Figure~\ref{fig:remedy_output}).

To facilitate continuous interaction, the application allows users to ask follow-up questions, such as the cause of a particular disease, alternative organic treatment options, or prevention techniques. The conversation is managed by LangChain’s memory component, which maintains context across multiple exchanges and provides relevant, reference-based answers (Figure~\ref{fig:followup_qa}). This makes the system not only a detection tool but also an intelligent assistant for agricultural guidance and decision-making.

\section{Conclusions}\label{Conclusion}

In this work, we presented an intelligent, end-to-end system that combines computer vision and natural language processing to assist in plant disease diagnosis and remediation. By fine-tuning the YOLOv8 model on an expert-curated dataset, we achieved reliable detection of major coffee leaf diseases. The integration of a Retrieval-Augmented Generation (RAG) framework enabled the system to retrieve relevant information from a domain-specific knowledge base, ensuring that the language model provided accurate, reference-backed remedies and explanations. 

Furthermore, the conversational interface built using Streamlit and LangChain memory enhanced user engagement by allowing follow-up questions and maintaining contextual understanding. This made the application not just a diagnostic tool, but a virtual assistant for agricultural decision-making. 

There is considerable scope for enhancing this system. One key area is improving the object detection component by upgrading to newer versions such as YOLOv10 or YOLOv11, which are expected to offer more accurate and efficient disease detection. Additionally, enriching the knowledge base with insights from domain experts can help generate more precise and contextually grounded responses. Importantly, the proposed framework is highly adaptable and can be extended to other agricultural domains or even entirely different fields where similar detection and explanation workflows are needed.

\section*{Acknowledgment}
We would like to express our gratitude to the Ultralytics platform for providing state-of-the-art object detection models that are easy to use and highly effective. We also sincerely thank the authors of related research works, whose contributions inspired our approach and provided valuable insights into both the conceptual framework and implementation details.

\bibliography{sn-bibliography}

\end{document}